\documentclass{article}
\usepackage{spconf,amsmath,graphicx}
\usepackage{hyperref}
\usepackage{comment}
\usepackage{tabularx}
\usepackage{todonotes}
\usepackage{paralist}
\usepackage{enumitem}

\usepackage{inputenc}
\usepackage{amsmath}

\title{SPARK: SPAcecraft Recognition leveraging Knowledge of Space Environment}
\name{\begin{tabular}{c}Mohamed Adel Musallam\textsuperscript{1}, Kassem Al Ismaeil\textsuperscript{1}, Oyebade Oyedotun\textsuperscript{1}\\
Marcos Damian Perez\textsuperscript{2} Michel Poucet\textsuperscript{2}, Djamila Aouada\textsuperscript{1}\thanks{This work was funded  by  the Luxembourg National  Research  Fund (FNR), under the project reference BRIDGES2020/IS/14755859/MEET-A/Aouada, and by LMO (https://www.lmo.space).}\end{tabular}}

\address{\textsuperscript{1} SnT, University of Luxembourg, \textsuperscript{2} LMO}

\begin{document}
\maketitle
\begin{abstract}
This paper proposes the SPARK dataset as a new unique space object multi-modal image dataset. 
Image-based object recognition is an important component of Space Situational Awareness, especially for applications such as on-orbit servicing, active debris removal, and satellite formation. However, the lack of sufficient annotated space data has limited research efforts in developing data-driven spacecraft recognition approaches. The SPARK dataset has been generated under a realistic space simulation environment, with a large diversity in sensing conditions for different orbital scenarios. It provides about 150k images per modality, RGB and depth, and 11 classes for spacecrafts and debris. This dataset offers an opportunity to benchmark and further develop object recognition, classification and detection algorithms, as well as multi-modal RGB-Depth approaches under space sensing conditions.
Preliminary experimental evaluation validates the relevance of the data, and highlights interesting challenging scenarios specific to the space environment. 
\begin{figure}[!t]
    \centering
    \includegraphics[width=.99\linewidth ]{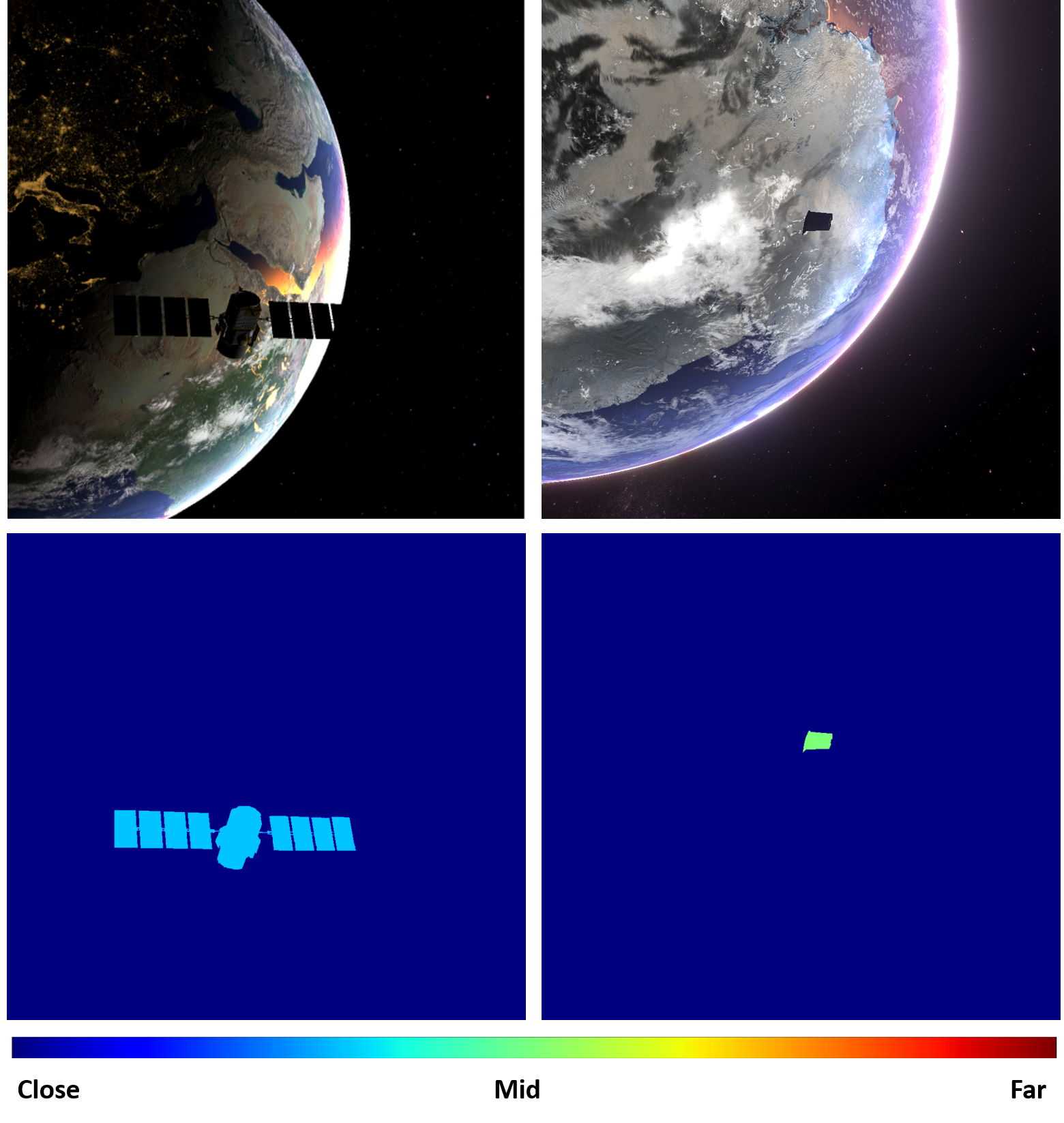}
    \caption{Samples from our \emph{SPARK} dataset. Top-left: RGB image of the \emph{`Calipso'} satellite with night side of Earth in the background. Top-right: RGB image of a debris with day side of Earth in the background. Bottom: corresponding depths.}
    \label{fig:CALIPSO}
    \vspace{-0.5 cm}
\end{figure}
\end{abstract}
\section{Introduction}
\label{sec:intro}
Today, our daily life is highly dependent on the increasingly growing satellite infrastructure. This infrastructure is used in many services; from communication, transportation, to weather forecast and many more. It is therefore essential to protect all space assets. One of the major threats is the risk of collision in space. Providing the spacecraft with the ability to autonomously recognize the surrounding objects is of utmost importance to minimize such risk and is one of the main objectives of Space Situational Awareness (SSA). 
Objects in question include active and inactive satellites, as well as space debris.
Over the past years, image-based sensors have been considered as a great source of information for SSA. This has triggered multiple research efforts in the field \cite{opromolla2017review,strube2015raven, yol:hal-01304728,chabot:hal-01784234,forshaw2016removedebris}, and recently, multiple works have been proposed to investigate the potential of deep learning (DL) from images for space applications~\cite{sharma2018pose,proencca2020deep}. \\ 
The performance of DL methods is highly dependent on the availability and quality of data used for training them. However, in the space domain, data are very scarce and costly to obtain. Efforts have been initiated, nonetheless, for the problem of 6D spacecraft pose estimation and two dedicated synthetic (or mixed with laboratory-acquired data) datasets were proposed for this purpose, i.e., the \textit{Spacecraft pose estimation dataset (SPEED)} ~\cite{speedchallange}~\cite{challenge} and the \textit{Unreal Rendered Spacecraft On-Orbit (URSO)}  dataset~\cite{proencca2020deep}.
However, to the best of our knowledge, no data exist so far for the task of space target recognition and detection while these tasks are extremely important for SSA and a crucial step towards reaching autonomy in space. Moreover, in view of the major advances made in object recognition using DL~\cite{liu2020deep}, it is interesting to investigate the applicability of these methods for space data, and identify directions for furthering their performance in the space environment.  

In this paper, we introduce a new dataset dedicated for the task of space target recognition and detection. This dataset, named \emph{SPAcecraft Recognition leveraging Knowledge of Space Environment (SPARK)} dataset, is a unique space multi-modal annotated image dataset. It contains a total of 150k RGB images with bounding box annotation for the target object in each image, and the same number, 150k, of depth images of 11 object classes, with 10 spacecrafts and one class of space debris. Sample images are shown in Fig.\ref{fig:CALIPSO}. The data have been generated under a photo-realistic space simulation environment, with a large diversity in sensing conditions, including extreme and challenging ones. The SPARK dataset will be shared publicly with the research community in the context of a competition\footnote{A corresponding challenge will be held during ICIP 2021 \url{https://cvi2.uni.lu/about-spark-2021/}}.
In addition to describing the different features of the SPARK data and their generation, we conduct experiments testing state-of-the-art (SoA) target recognition methods, highlighting interesting challenges for the research community. These include looking into the important question of multi-modal RGB-Depth space object recognition with preliminary results encouraging the added value of multi-modal learning.

The rest of the paper is organized as follows: Section~\ref{sec:Related work} describes related SSA datasets. Details about the simulator and the proposed SPARK dataset are given in Section~\ref{SPARK}. Section~\ref{sec:results} presents the conducted experiments and discusses interesting uses of the proposed data. Section~\ref{sec:conclusion} concludes this work. 

\section{Related Work} \label{sec:Related work}
 There are multiple datasets available for the task of object recognition. Most of them contain real images~\cite{ILSVRC15,lin2014microsoft,Geiger2013IJRR,DBLP:journals/corr/abs-1803-06184}, while some contain synthetically-generated images~\cite{sadat2018effective,cabon2020virtual,Ros_2016_CVPR}. In the space domain, given the difficulty of obtaining large real datasets, synthetic datasets are currently the default approach for developing DL methods for SSA tasks. Moreover, we have identified two such datasets only~\cite{proencca2020deep,speedchallange}, and both have been designed specifically for spacecraft 6D pose estimation.  We herein review and discuss these datasets, comparing them to the SPARK dataset as summarized in Table~\ref{tab:Table1}. 

SPEED \cite{speedchallange} consists of 15,300 gray scale images of the same mock-up spacecraft, with a resolution of 1920×1200 pixels. Two types of images are used: a) 300 images are lab captured. The lab setup employs a robotic arm to position and orient optical sensors with respect to the target mock-up, custom illumination devices to simulate Earth albedo, and Sun simulator to emulate the illumination conditions present in space. b) 15k images of the same mock-up are synthetically generated using an augmented reality software that fuses synthetic and actual space imagery. The testing set contains 300 real lab images and $\sim$3k synthetic images, whereas the training set contains 12k synthetic images and only 5 real lab images. The ground truth 6D pose of the mock-up with respect to the optical sensor is calculated using a motion capture system with calibrated cameras.

URSO~\cite{proencca2020deep} provides three datasets of satellite images at a resolution of 1080×960 pixels. One dataset is for the \emph{`Dragon'} spacecraft and two datasets for the \emph{`Soyuz'} spacecraft with different operating ranges. Each of these datasets contains 5k images, of which 10\% are dedicated for testing and 10\% for validation. The images were randomly generated and sampled around the day side of the Earth from low Earth orbit altitude. The Earth rotation, camera orientation, and target object pose are all randomized. The target object is placed randomly within the camera field of view and an operating range between 10m and 40m. All images are labelled with the corresponding target pose with respect to the virtual vision sensor.

While the two datasets, SPEED and URSO, are dedicated for the task of 6D pose estimation of a satellite, the proposed SPARK dataset is designed for a different SSA task; that of space object detection, including \emph{localization, segmentation and classification}. Indeed, our dataset covers a larger number of satellite models and space debris forming \emph{11 object classes} with their annotated 2D images, depth images, and segmentation masks. This is detailed in Section~\ref{SPARK} below.  

\begin{table}[t!]
    \centering
    \vspace{-0.8 cm}
    \begin{tabular}{l}
         \includegraphics[width=\linewidth]{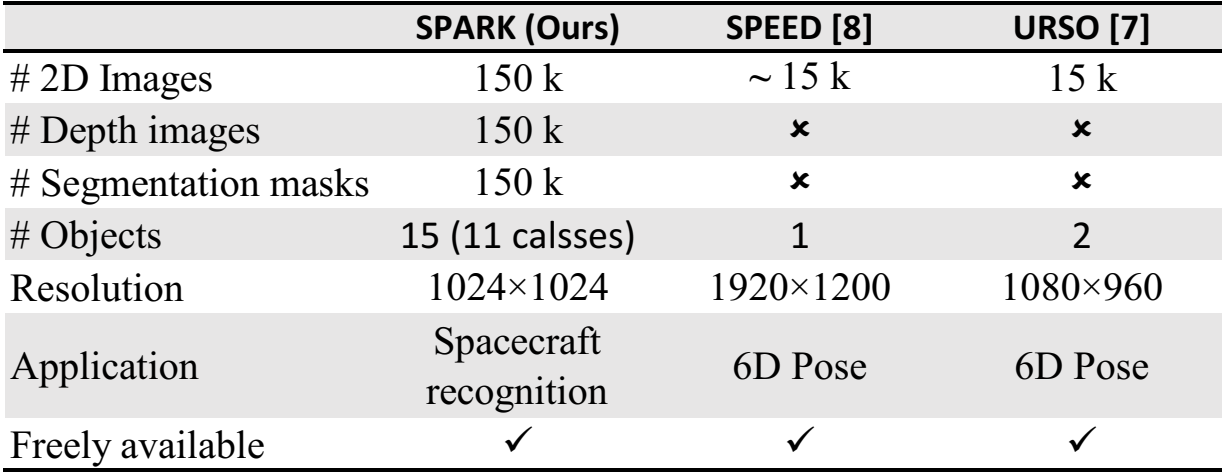}
    \end{tabular}
    \caption{Comparison of SSA related learning datasets.}
    \label{tab:Table1}
     \vspace{-0.5 cm}
\end{table}

\section{Proposed SPARK dataset} \label{SPARK}
The proposed SPARK dataset was generated using the Unity3D game engine as a simulation environment~\cite{unity}. The simulation consists of the following main components:\\
\textbf{Earth model:} It consists of a high resolution textured and realistic Earth model composed of 16k polygons, obtained from a third party Unity asset \cite{Asset}, based on the NASA Blue Marble collection~\cite{blue-marble}. It presents clouds, clouds' shadows, and atmospheric outer scattering effects. This model is located at the center of the 3D environment with respect to all orbital scenarios. The surrounding space background uses a high resolution panoramic photo of the Milky way galaxy from the European Southern Observatory (ESO)\cite{eso}. \\
\textbf{Target spacecraft:} Ten different realistic models of spacecrafts were used (\textit{`AcrimSat', `Aquarius', `Aura', `Calipso', `CloudSat', `Jason', `Terra', `TRMM', `Sentinel-6'}, and the \textit{`1RU Generic CubeSat'}). They were obtained from NASA 3D resources~\cite{nasa}. The debris are parts of satellites and rockets after adding corrupted texture to simulate dysfunction conditions (\emph{`space shuttle external tank', `orbital docking system', `damaged communication dish', `thermal protection tiles',} and \textit{`connector ring'}). They were placed around the Earth and within the low Earth orbit (LEO) altitude.\\ 
\textbf{Chaser spacecraft:} It represents the observer equipped with different vision sensors used to acquire data.\\
\textbf{Camera:} A pinhole camera model was used with known intrinsic camera parameters and optical sensor specifications, as well as a depth camera for range imaging.

\begin{figure}[t!]
\includegraphics[width=\linewidth , height=3.5cm]{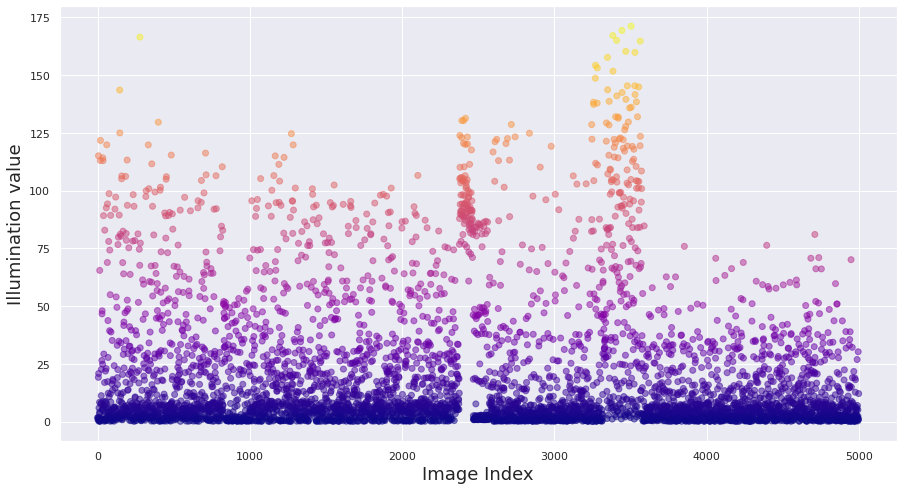} 
\includegraphics[width=\linewidth , height=3.5cm]{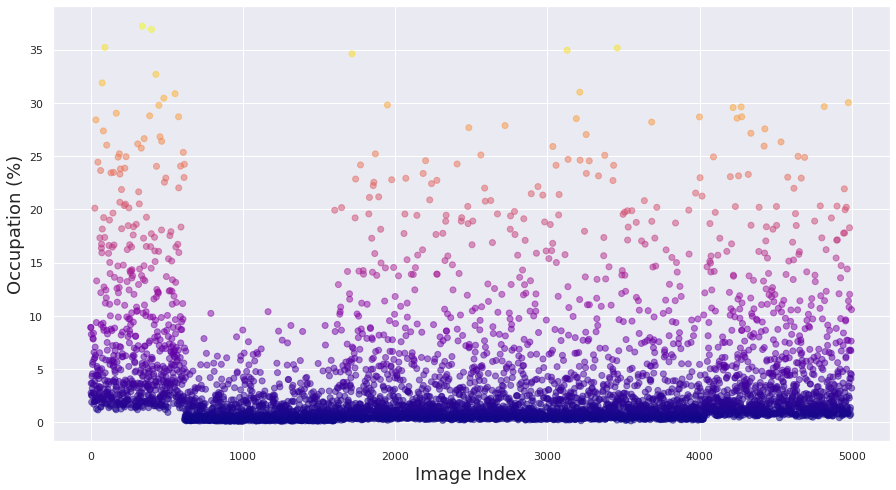}
\caption{Data distribution of the \emph{`Sentinel-6'} satellite with respect to:  illumination (top) and range or target occupation of the scene (bottom).}
\vspace{-0.5 cm}
\label{fig:cubesatIllumination}
\end{figure}

The SPARK dataset is generated by randomly placing the target satellite or spacecraft within the field of view of a camera mounted on a chaser. We consider different ranges and orientations of the chaser model. Furthermore, the Sun and Earth are randomly rotated around their respective  axes in every frame. This allows to obtain high-resolution photo-realistic RGB images, depth maps, and the corresponding segmentation masks in multiple and different environmental conditions. The quality of spaceborne imaging is highly dependent on many factors such as varying illumination conditions, signal-to-noise ratio, and high contrast. Therefore, the SPARK dataset has been created in a way that covers a wide range of cases, including extreme and challenging ones. The whole dataset is spanned by sampling the following axes:\\
\textbf{Scene illumination:} We model the prominence of the Sun flares, rays, and reflections on Earth from the space in order to represent the different illumination conditions and the high contrast of the spaceborne images. Extreme illumination cases are considered where the sunlight directly faces the optical navigation sensors or reflects from the target surface or the Earth, hence causing lens flare and optical sensor blooming effect (see example distribution in Fig.~\ref{fig:cubesatIllumination} (top)).\\
\textbf{Scene background:} In different orbital scenarios, the target spacecraft can be oriented towards the Earth or towards a dark space, which means that the target spacecraft's background can change with respect to different positions and orientations. When the Earth is in the background, the scene will have additional features of the planet's surface and high reflectively of oceans and clouds. Whereas, having a dark space in the background will lead to featureless space with sparsely illuminated stars in the background.\\ 
\textbf{Distance between the camera and the target:} We model different ranges between the target spacecraft and the optical sensor mounted on the chaser spacecraft. Range is inversely proportional to the percentage of target occupation of the scene (see example distribution in Fig.~\ref{fig:cubesatIllumination} (bottom)). \\
\textbf{Optical sensor noise:} Spaceborne images suffer from high noise levels due to small sensor sizes and high dynamic range imaging \cite{sharma2018pose}. Accordingly, varying levels of zero-mean white Gaussian noise were added to the generated synthetic images to simulate the noise in real spaceborne images.

The final SPARK dataset consists of $\sim$150k high-resolution photo-realistic RGB images with bounding box annotation for the target object in each image, $\sim$150k depth images, and the corresponding $\sim$150k segmentation masks in multiple and different space environmental conditions. It includes 10 different satellites, 12.5k images for each satellite, and 5 debris objects, with 5k images for each one, and all debris combined in one class.

\begin{figure*}[!th]
    \centering
    \begin{minipage}{.33\textwidth}
        \centering
        \includegraphics[width=\linewidth,height=3.5cm]{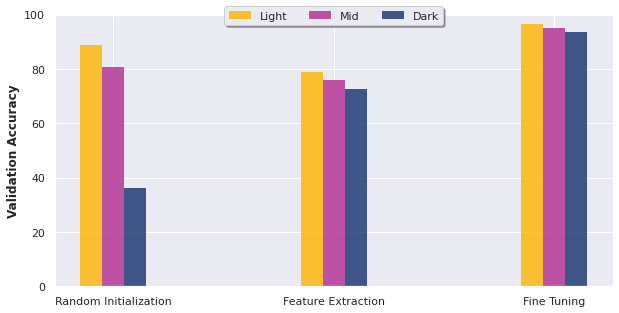}

        (a)
    \end{minipage}%
    \begin{minipage}{0.33\textwidth}
        \centering
        \includegraphics[width=\linewidth,height=3.5cm]{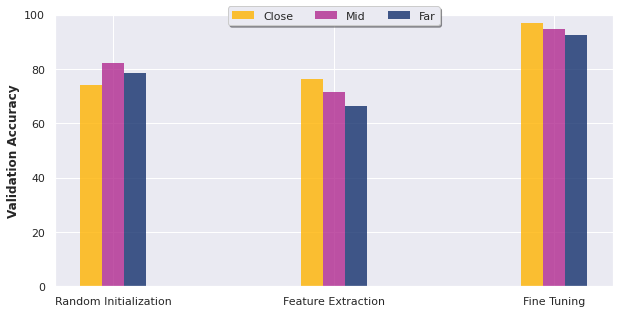}

        (b)
    \end{minipage}
    \begin{minipage}{0.33\textwidth}
        \centering
        \includegraphics[width=\linewidth,height=3.5cm]{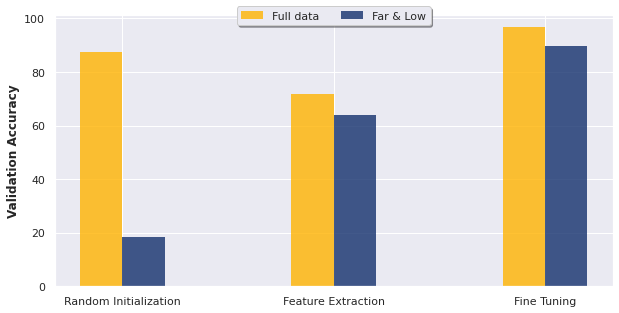} 

        (c)
    \end{minipage}
    \caption{Accuracy of ResNet18 considering the three cases of \textbf{\texttt{Random Initialization}}, \textbf{\texttt{Feature Extraction}}, and \textbf{\texttt{Fine Tuning}}. (a) Different illumination levels (High, Mid, and Low). (b) Different ranges (Close, Mid, and Far). (c) Training using full dataset versus training using challenging cases containing far range with low illumination images .}
        \label{fig:plot_results_illumination_range}
        \vspace{-0.5 cm}

\end{figure*}

\section{Experiments} \label{sec:results}

In order to analyze the features of the SPARK dataset and highlight interesting research questions this novel dataset opens, we conduct multiple space object classification experiments using SoA approaches. 
We run two categories of experiments to look into two aspects: 1) impact of the scene and sensor, analyzing the effect of scene illumination, object occupation and noise contamination; and 2) potential impact of multi-modal spacecraft recognition.\\
\textbf{Impact of scene and sensor:}
We tested different variants of ResNet~\cite{He_2016_CVPR} and EfficientNet~\cite{tan2020efficientnet}. Specifically, we tested three different models, i.e., ResNet18, ResNet34, and EfficientNet B0, and for each model we considered three cases: \\
(a) \textbf{\texttt{Random Initialization}}: where training is done from scratch without any pre-trained weights, with a classifier on top of the feature extraction backbone; \\
(b) \textbf{\texttt{Feature Extraction}}: where we apply transfer learning on the model pre-trained on ImageNet~\cite{ILSVRC15} then freeze the feature extraction and train a classifier on top of it; \\
(c) \textbf{\texttt{Fine Tuning}}: where we use a model pre-trained on ImageNet and add a classifier on top of it. Then, we retrain both on the dataset under consideration.
\\
All the tested networks were trained using ADAM optimizer with a $1e-4$ learning rate and a batch size of 11 images, with cross entropy loss for classification. For ResNet18 and EfficientNet B0, we trained for 20 epochs and for ResNet34 we trained for 50 epochs. Datasets have been split as follows, 80\% for training and 20\% for validation. For practicality, all images were resized to 512 by 512 pixels.

\begin{figure}
\includegraphics[width=\linewidth , height=3.5cm]{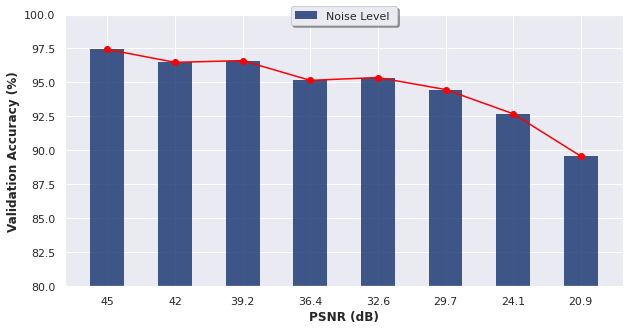} 
\caption{Validation accuracy of a fine-tuned ResNet18 with respect to the average peak signal-to-noise ratio (PSNR) due to the added Gaussian noise.}
\label{fig:NosieLevel}
\vspace{-0.5 cm}
\end{figure}


\noindent We found that scene illumination level and range have direct impact on the performance of DL algorithms regardless of the model architecture\footnote{We plot the results for ResNet18, but similar results have been obtained using ResNet34 and EfficientNet B0.}. For instance, we can see in Fig.\ref{fig:plot_results_illumination_range}(a) that training using subset of the dataset with high illumination leads to higher validation accuracy as compared to mid and low illumination levels regardless of the tested models and initialization schemes. Similar behavior was observed while training with images with different ranges as reported in Fig.\ref{fig:plot_results_illumination_range}(b) where the accuracy generally drops by increasing the distance between the camera and target object. \\
In order to address extreme cases, we considered a subset of our dataset that contains only images with low illumination and far ranges. This is considered to be a very challenging scenario for SSA tasks using a monocular camera. We observed a high drop in the accuracy in the case of \textbf{\texttt{Random Initialization}}, and a drop of around $ 7\%$ for the two other initialization schemes (see Fig.\ref{fig:plot_results_illumination_range}(c)). \\
Overall, we found that the \textbf{\texttt{Fine Tuning}} initialization scheme provides the best results across all the conducted experiments, while using the \textbf{\texttt{Feature Extraction}} scheme leads to lower performance as compared to \textbf{\texttt{Random Initialization}} (see Fig.\ref{fig:plot_results_illumination_range} (a) and (b)). \\
Finally, we tested the impact of the sensor noise on classification accuracy by progressively increasing the level of contamination with noise We observe  a systematic decrease in performance as reported in Fig.\ref{fig:NosieLevel}. \\
\textbf{Multi-modal spacecraft recognition:}
In order to show the potential of multi-modal spacecraft classification, we tested the multi-modal classification approach proposed in~\cite{oyedotun2019learning}. We retrain from scratch on the SPARK dataset using both the provided RGB and depth data modalities. The validation accuracy was boosted to 90.05\% when fusing data as opposed to 75\% and 88.01\% when learning respectively from RGB data, or depth data alone. 
Note that these preliminary results have been obtained using largely resized input images, i.e., 64 by 64 pixels, during training. The objective of this experiment is mainly to show the advantage that multi-modal data gives to the task of SSA. A more thorough evaluation with larger image resolutions will be reported in a dedicated paper.

\section{Conclusion} \label{sec:conclusion}
We proposed the SPARK dataset for spacecraft recognition, localization, and segmentation. This dataset was generated under a realistic space simulation environment, providing a large range of diversity in sensing conditions and multiple spacecraft classes. 
This dataset is proposed to pave the way for developing dedicated deep learning approaches for debris and spacecraft recognition and detection in the context of SSA missions. 
First experiments on SPARK identified the most challenging sensing conditions that are important to focus on, i.e., far range, low illumination, and large contamination with noise. Furthermore, multi-modal RGB-Depth spacecraft recognition is identified as a relevant research direction. A competition using the SPARK dataset will follow to foster efforts on data-driven SSA.

\label{sec:ref}

\bibliographystyle{IEEEbib}
\bibliography{refs}

\end{document}